\documentclass{article}

\usepackage[english]{babel}

\usepackage[letterpaper,top=2cm,bottom=2cm,left=3cm,right=3cm,marginparwidth=1.75cm]{geometry}

\usepackage{comment}
\usepackage{amsmath}
\usepackage{amssymb}
\usepackage{authblk}
\usepackage{graphicx}
\usepackage[colorlinks=true, allcolors=blue]{hyperref}

\title{Developing Explainable Machine Learning Model using Augmented Concept Activation Vector}

\author[1,2]{Reza Hassanpour\thanks{Corresponding Author: r.zare.hassanpour@rug.nl}}
\author[2]{Kasim Oztoprak\thanks{kasim.oztoprak@gidatarim.edu.tr}}
\author[3]{Niels Netten\thanks{c.p.m.netten@hr.nl}}
\author[3]{Tony Busker\thanks{a.l.j.busker@hr.nl}}
\author[4]{Mortaza S. Bargh\thanks{m.shoae.bargh@wodc.n}}
\author[3,4]{Sunil Choenni\thanks{r.choennie@hr.nl}}
\author[2]{Beyza Kizildag\thanks{beyzaberen01@gmail.com}}
\author[2]{Leyla Sena Kilinc\thanks{leylasenakilinc@gmail.com}}
\affil[1]{Computer Science Department, Groningen University, the Netherlands}
\affil[2]{Computer Engineering Department, Konya Food and Agriculture University, Turkey}
\affil[3]{Research Center Creating 010, Rotterdam University, the Netherlands}
\affil[4]{Research and Data Center, Ministry of Justice and Security, The Netherlands}

\begin{document}
\maketitle

\begin{abstract}
Machine learning models use high-dimensional feature spaces to map their inputs to the corresponding class labels. However, these features often do not have a one-to-one correspondence with physical concepts understandable by humans, which hinders the ability to provide a meaningful explanation for the decisions made by these models. We propose a method for measuring the correlation between high-level concepts and the decisions made by a machine learning model. Our method can isolate the impact of a given high-level concept and accurately measure it quantitatively. Additionally, this study aims to determine the prevalence of frequent patterns in machine learning models, which often occur in imbalanced datasets. We have successfully applied the proposed method to fundus images and managed to quantitatively measure the impact of radiomic patterns on the model's decisions.
\end{abstract}

\section{Introduction}

Machine learning models have achieved remarkable accuracy in decision-making. However, their complexity makes it difficult for humans to comprehend their inference process. In most cases, machine learning models are evaluated based on the correlation between their input and output. While these black-box models may seem sufficiently acceptable, there are scenarios where transparency in the decision-making process is essential to certify the model's validity and clarify the potential risks associated with the suggested decisions.

An important application area where such transparency plays a crucial role is medicine. In this field, both practitioners and patients are keen to consider the safety of the decisions made by machine learning models. Consequently, many researchers consider transparency to be an important factor in making the model trustworthy. The added transparency helps explain the factors that influence the selection of an output and its relationships with contextual parameters.

Therefore, explainability should be addressed in the context of the model's application. Subsequently, the transparent model can be improved, verified, and, in some cases, reveal hidden characteristics of the data. An important challenge in developing explainable models is the lack of explicit correlations between low-level features extracted by deep learning models and human-understandable concepts, such as radiomics used in medical applications. This issue implies that even explaining the inference process of a model may not be easily comprehensible to humans.

The second challenge is the importance attached to specific features by domain experts, even though these features may be less frequent or relatively rare. In this article, we address both problems. We propose a novel method for establishing a correlation between human-understandable concepts and low-level model parameters. Additionally, we consider the impact of imbalanced data on machine learning models. We have used fundus images of patients with retinopathy complications to validate our proposed model.

This article is structured as follows. In Section \ref{sec1}, we review the related literature. Section \ref{sec2} details the proposed model. In Section \ref{sec3}, we present the experimental results of our research, and finally, in Section \ref{sec4}, we draw our conclusions and discuss the results.

\section{Related Works}
\label{sec1}
Deep learning techniques have demonstrated remarkable efficacy across a range of medical diagnostic tasks, even surpassing human experts in some instances \cite{zeiler2014visualizing}. Nonetheless, the opacity inherent to these algorithms has limited their practical clinical adoption\cite{marcinkevivcs2023interpretable}. Recent investigations into explainability strive to reveal the primary drivers behind a model's decisions. The impetus behind these inquiries arises from the realization that elucidation plays a pivotal role in critical medical treatments like cancer care. In such applications, the importance of predictions extends beyond their precision; their comprehensibility also holds significant weight. Yet, concurrently achieving both attributes poses a formidable challenge, as there exists a delicate balance between interpretability and accuracy \cite{rizopoulos2018book}.

Explainable AI (XAI) in medical image processing focuses on developing techniques that enhance the interpretability and transparency of artificial intelligence models used in medical imaging \cite{abbasi2017structural}. This is crucial for gaining trust from medical practitioners, ensuring patient safety, and providing insights into the decision-making process of AI systems.

In the realm of machine learning literature, explainability techniques are classified into two main categories: model-based \cite{tsang2017detecting}, \cite{selvaraju2017grad} and post-hoc \cite{springenberg2014striving}, \cite{tu2020sunet} explanations\cite{mitu2023explainable}. When delving into a trained model to gain a deeper understanding of the acquired correlations, this approach is termed post-hoc explanation. A crucial differentiation between post-hoc and model-based explanations lies in their methodologies. The former involves training a neural network and then seeking to explicate the operations of the resulting black-box network, while the latter mandates that the model itself is inherently interpretable.

Researchers have considered explainability from different perspectives. Zhou et al. proposed an architecture to provide more interpretable results. They claim that their model is based on networks with attention mechanisms or gradient-based visualization methods that can highlight regions of interest in medical images, making the predictions more understandable \cite{zhou2016learning}.
In a similar approach, Simonyan et al. proposed a heatmap or saliency map model. Their model uses techniques that highlight the regions of input images most influential in making a prediction, aiding in understanding why the AI model arrived at a particular decision \cite{simonyan2013deep}. Techniques like LIME (Local Interpretable Model-agnostic Explanations) and SHAP (SHapley Additive exPlanations) provide post-hoc explanations for predictions from any machine learning model, enhancing interpretability in the medical imaging domain \cite{ribeiro2016should}.

LIME (Local Interpretable Model-agnostic Explanations)\cite{palatnik2019local} and SHAP (SHapley Additive exPlanations)\cite{nohara2022explanation} are two popular techniques for explaining the predictions of machine learning models. Both techniques are model-agnostic meaning that they can explain any type of model and locally interpret individual predictions. LIME approximates the model locally around a specific prediction by training a simple, interpretable model. This approximation is based on perturbing input data and observing changes in predictions to create a linear, interpretable model around a specific instance. As the result, LIME provides weights of features in the local linear model, making it possible to interpret the impact of each feature on the decision of the model. SHAP in addition outputs Shapley values, which represent each feature's contribution to the prediction in a fair way. 

On the other hand, some researchers have focused on making the model consider specific features. Attention mechanisms in deep learning models enable the model to focus on particular parts of the image, contributing to a more transparent decision-making process. These mechanisms have been used to improve performance and interpretability in medical image analysis \cite{oktay2018attention}.

Radiomics involves extracting a large number of quantitative features from medical images. Explainable modeling techniques applied to radiomics help identify which features are most influential in making a particular diagnosis or prediction \cite{kumar2012radiomics}. These techniques are also used to reduce the dimensionality of the feature space, emphasizing the most salient features \cite{hassanpour2023adaptive}. Explainable radiomics involves applying interpretable techniques to the process of extracting quantitative features from medical images, such as CT scans, MRI images, and X-rays, to gain insights into the factors contributing to a diagnosis or prediction.

Kumar et al., in their foundational paper, introduce the concept of radiomics and discuss the process of extracting a large number of quantitative features from medical images. They highlight the challenges in radiomics, including feature selection and validation \cite{kumar2012radiomics}. Lambin et al. discussed the potential of radiomics to extract more information from medical images using advanced feature analysis techniques and emphasized the need for robust and reproducible radiomic features \cite{lambin2012radiomics}.

Aerts et al. demonstrated how radiomics can be used to decode tumor phenotypes by analyzing noninvasive medical images, showcasing the potential of radiomics to provide insights into tumor characteristics \cite{aerts2014decoding}. In their research, Parmar et al. applied machine learning to radiomic features extracted from head and neck cancer images to develop prognostic biomarkers, demonstrating the potential of radiomics in predictive modeling \cite{parmar2015radiomic}. Nie et al. focused on breast MRI and demonstrated how quantitative analysis of lesion morphology and texture features can aid in diagnostic prediction, highlighting the potential of radiomics in characterizing breast lesions \cite{nie2008quantitative}. Zhang et al. introduced the IBEX software platform, which is designed to facilitate collaborative work in radiomics, emphasizing the importance of standardized tools for radiomic feature extraction and analysis \cite{zhang2015ibex}.

The Testing Concept Activation Vector (TCAV) method proposed by Kim et al. is an interpretability technique designed to assess the influence of high-level concepts on the predictions made by machine learning models \cite{kim2018interpretability}. TCAV works by comparing the model's responses when the concept of interest is present versus when it is absent. By manipulating the concept while keeping other factors constant, TCAV calculates a concept's impact on the model's predictions using statistical tests. If the model's predictions are found to be sensitive to the concept, it suggests that the model has learned to associate the concept with the prediction, thereby providing insights into how the model uses the concept to make decisions. In essence, TCAV provides a way to understand and measure the extent to which specific high-level concepts are used by a machine learning model to arrive at its predictions. This method has applications in various domains, including medical image analysis, where it can help ensure that models are making decisions based on clinically relevant factors.

 Our proposed method also uses high-level concepts to determine to what extent decisions made by the model are affected by those high-level patterns. However, our method distinguishes itself from the TCAV method by augmenting input data with concept patterns and restricting the impact of external factors. Additionally, our method uses a single network for both training-classification and explainability verification. In contrast to both SHAP and LIME techniques, the input to the our prposed model consists of raw data rather than extracted features, which is processed by a deep neural network. The proposed model therefore distinguishes itself by establishing a correspondence between the model's internal features and the high-level radiomic patterns used by practitioners.

\section{Augmented Concept Activation Vector}
\label{sec2}
We propose a new method for evaluating the impact of human-understandable visual patterns on the classification decisions of machine learning models. Our proposed method follows the same idea as the Testing Concept Activation Vector (TCAV) in measuring the contribution of a concept (high-level visual pattern) to the decisions of machine learning models. However, it distinguishes itself from TCAV in a few ways. Firstly, our proposed method called the Augmented Concept Activation Vector (ACAV) method utilizes the context of its input data to minimize the effect of external factors.

 While TCAV considers two sets of data points $P_C$ and N to represent the positive and the negative samples of the given concept, the ACAV defines data point sets $T_P$ and $T_N$ where $d^n\subset \mathbb{R}^n$ is the input data points set, $T_P\subset d^n$ and $T_N\subset d^n$ are subsets of the original dataset. $T_P$ includes the data points with the given concept pattern and classified accordingly, while data points in $T_N$ are samples without having the concept pattern that are classified correctly. 

In addition $A_C:\{a_C | a_C=a\bigoplus C,a\in T_N,C\space is\space a\space concept\space pattern\}$  a subset of $T_N$ data points, is defined to represent the date points with an augmented concept. The main advantage of augmenting the concept pattern to a data point is preserving the context of the sample and avoiding external factors. The ACAV therefore targets application areas where the input data points share a strong context such as medical images. 

\begin{figure}
\centering
\includegraphics[width=.75\linewidth]{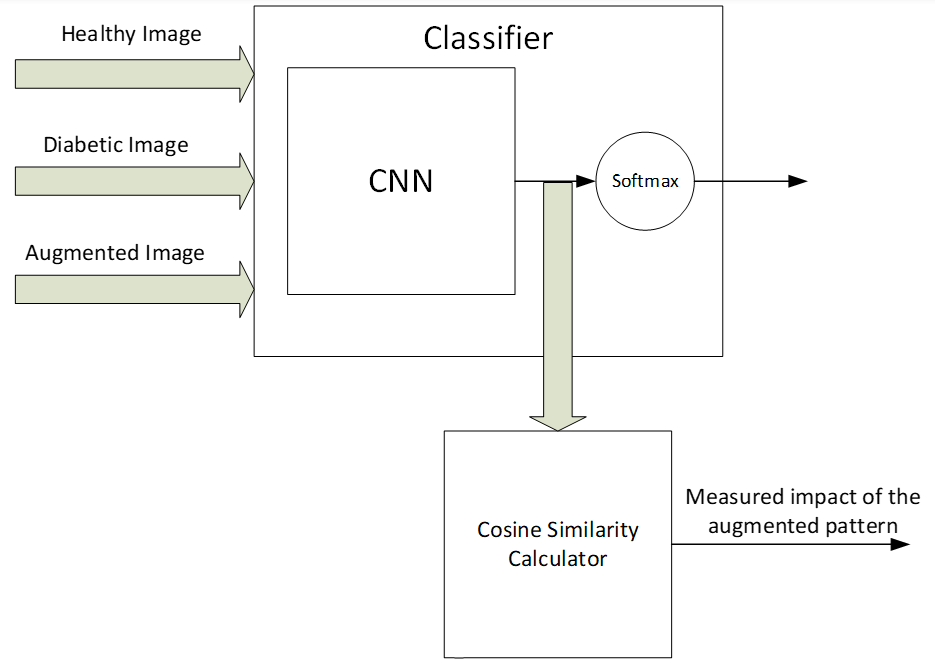}
\caption{\label{fig:diagram}Schematic diagram of the proposed method.}
\end{figure}

Secondly, ACAV can handle imbalanced datasets where certain concept patterns occur less frequently but are highly valued by domain experts for classification purposes (e.g., rare symptoms used for more reliable diagnoses in medical applications). Finally, ACAV employs a single machine learning model, providing greater flexibility in designing and implementing the method as shown in Figure \ref{fig:diagram}.

We propose training a machine learning model to classify data points as belonging to one the possible classes. Our assumption is that the classification process considers the existence and prevalence specific concept patterns. The activation vector of layer \textit{l} maps an input vector of dimension \textit{n} to a vector of dimension \textit{m}$\ll$\textit{n} as $f_l: \mathbb{R}^n\rightarrow \mathbb{R}^m$. The direction of the activation vector alters with the value of the input data, however, the differences are limited when the input points belong to the same class (and classified correctly).  Equation \ref{eq:e1} defines the activation of neuron \textit{i} of layer \textit{l}.
\begin{equation}
\label{eq:e1}
   a_i^{(l)} = f\left(\sum_{j=1}^{n^{(l-1)}} w_{ij}^{(l)} a_j^{(l-1)} + b_i^{(l)}\right) 
\end{equation}
The activation vector of layer l is defined as shown in Equation \ref{eq:e2}.
\begin{equation}
    \label{eq:e2}
    \mathbf{a}^{(l)} = f\left(\mathbf{W}^{(l)} \mathbf{a}^{(l-1)} + \mathbf{b}^{(l)}\right)
\end{equation}
First, we train a model to classify the labeled input data. Subsequently, we consider a sample data point $d_i$ lacking the concept pattern C which is classified as $S_i$ by the model. The activation vector at layer \textit{l} when this data point is fed into the model is given as $V_\textit{l} = f(d_i)$. \\
Next, we augment the data point by including the concept C pattern as $d_{ia}=d_i\bigoplus C$. We give the augmented data point to the model and find the activation vector at layer \textit{l} as $V_{la} = f(d_i)$. The experiment is repeated with n data samples and the average rate of deviation of the activation vector is found using Equation \ref{eq:e3}  
\begin{equation}
    \label{eq:e3}
    \Delta V=  \frac{1}{n} \sum_i|V_{la}-V_l|
\end{equation}

To evaluate the impact of a given concept on the decision-making process of the model, we define two metrics. The first metric measures the relative deviation of the activation vectors when inputs are augmented with a concept pattern. The activation vectors of fully connected feedforward networks serve as the foundation for classification decisions made in the output layer. These vectors represent the combined processing of features extracted in earlier layers of the neural network, such as convolutional layers. Consequently, any changes in the input data are reflected in the activation vectors. Our goal is to measure deviations in the activation vectors when changes in the input data are exclusively due to the presence of high-level concepts. This metric is the ratio of the number of data points that are assigned to a different class after being augmented to the number of data points that preserve their initial class labels. This metric is an indication of the contribution of the visual pattern to the decision made by the model.
The second metric takes into account the abundance of visual concepts when more than one concept is present in the data points. If a concept appears more frequently in the data set, it becomes the dominant factor in determining the class attributes. Therefore, imbalance data can deteriorate the accuracy of the model.  We measure data imbalance using an entropy-based metric as shown in Equation \ref{eq:e4}.  
\begin{equation}
    \label{eq:e4}
    H = -\sum_{i=1}^C p_i \log(p_i)
\end{equation}
where \textit{H} is the entropy of the high-level pattern distribution, $p_i$ is the proportion of high-level patterns in class \textit{i}, and \textit{C} is the total number of high-level patterns. 
Our second metric aims at measuring how effective the distribution of different visual concepts is on the decisions of the model.
Figure \ref{fig:diagram} depicts a schematic diagram of our proposed method.
\section{Experimental Results}
\label{sec3}
To experimentally validate the proposed method, we have conducted three sets of experiments. The dataset used in our experiments is the publicly available dataset RFMiD\footnote{RFMiD dataset (RETINAL FUNDUS MULTI-DISEASE IMAGE DATASET).´Center of Excellence in Signal and Image Processing, Shri Guru Gobind Singhji Institute of Engineering and Technology, Nanded, India}  dataset. The dataset includes 3200 fundus images labeled with one of the five diabetics classes where class zero includes healthy cases, and classes one through four represent diabetics classes with different severity (four having the most acute cases). The images corresponding to the less severe cases of diabetics incorporate fewer occurrences of radiomic patterns and limited to only some of those patterns. Therefore, in our experiments we considered only two classes of healthy and diabetics cases. We consider images of the healthy cases as the training samples of the first class, and the fourth, and fifth categories as samples belonging to the second class.

\begin{figure}
\centering
\includegraphics[width=.75\linewidth]{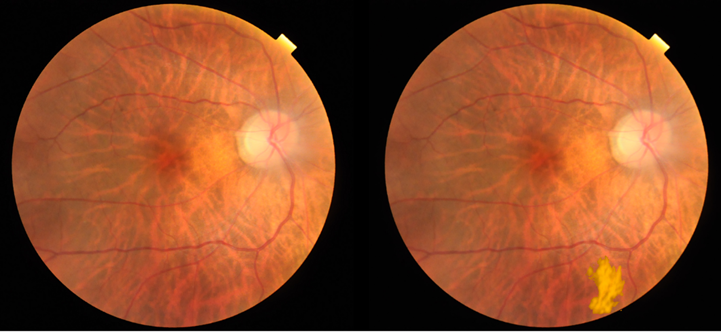}
\caption{\label{fig:healthy}Healthy images augmented with cotton-wool radiomic pattern.}
\end{figure}

\begin{figure}
\centering
\includegraphics[width=.75\linewidth]{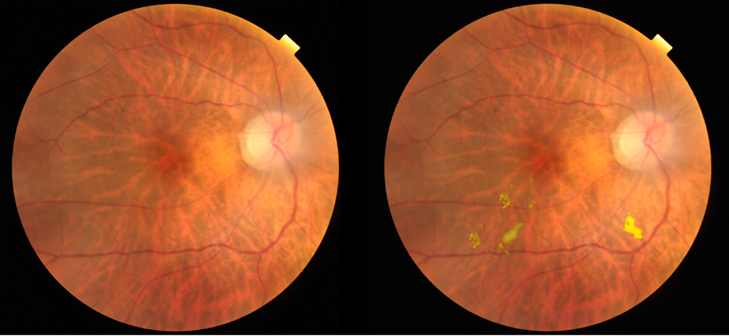}
\caption{\label{fig:aug1}Healthy images augmented with cotton-wool and fatty dots radiomic patterns.}
\end{figure}

\begin{figure}
\centering
\includegraphics[width=.75\linewidth]{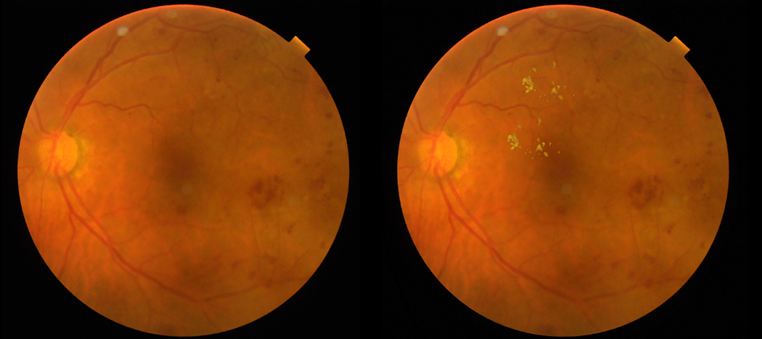}
\caption{\label{fig:aug2}Images with bleeding pattern augmented with fatty dots radiomic pattern.}
\end{figure}

A subset of 50 healthy fundus images have been augmented by adding a single cotton-wool pattern at locations close to blood vessels. A sample image of this subset is illustrated in Figure \ref{fig:healthy} where the original images and the augmented images are shown.
The second experiment includes more than one augmented pattern per image in a subset of 50 healthy images. Figure \ref{fig:aug1} illustrates a sample from this subset.
The third experiment aims at measuring the impact of a less frequent symptom such as cotton-wool patterns when the image includes multiple high-level features. This experiment is to determine the behavior of the model in presence of imbalanced data. For this experiment we have augmented images having bleeding symptom with fatty dots pattern. As mentioned, category two and three have not been considered as part of the experiments. However, some of these images are used to augment them with less frequent radiomic patterns as they generally include only more frequent bleeding pattern. A third subset with 50 images has been used for the third experiment. Figure \ref{fig:aug2} illustrates a sample augmented image used for the third experiment.

\begin{figure}
\centering
\includegraphics[width=.65\linewidth]{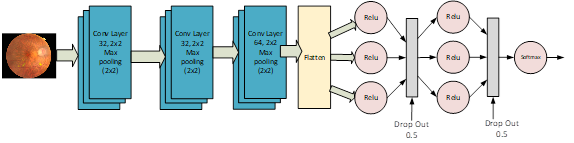}
\caption{\label{fig:schem}Schematic diagram of the classifying model.}
\end{figure}

We used a CNN with three convolution layers, and three layers at the fully connected MLP where the last layer includes five neurons having soft-max activation function. Figure \ref{fig:schem} shows a schematic structure of the model with parameters used.
The model is trained using two classes of training data, namely healthy and diabetics. Next using test data, we evaluate the performance of the network. While classifying healthy images, if an image is correctly classified, we store the values of the layer before the last as a vector of 64x1 dimension. At the end we find the average of the stored vectors. In a similar way we calculate the output of the layer before last for diabetic image that are classified correctly. These two vectors indicate the direction of the activation vector for two different classes. 
As shown in the diagram of Figure \ref{fig:schem}, the last layer includes a single neuron with softmax activation function. To avoid cases where the model is not very confident about, we use softmax neuron output values greater than 0.6 as healthy, and those values less than 0.4 as unhealthy. Hence, an error margin of 0.2 is considered.

Subsequently, we provided the augmented images as input and measured the activation vector value of layer before last. This vector is compared with the average activation vectors of the healthy and the diabetic inputs using cosine similarity. As the original images before augmenting radiomic patterns are available, we can find the deviation of the activation vector from its original direction.  Table \ref{tab:tab1} summarizes the results as explained.
\begin{table}
\centering
\begin{tabular}{|c|c|c|c|}
\hline
Augmented & Average Norm Vector & Average Norm Vector & Average Absolute \\
 Pattern Type &   Original Image &  Augmented Image  & Deviation \\\hline
Fatty dots  (single pattern)&	0.864	& 0.835 & 0.03\\
Fatty dots  (multiple patterns) & 0.864	& 0.813& 0.06 \\
Cotton wool   (single pattern) &	0.864&	0.831	&0.03 \\
Cotton wool  (multiple patterns) &	0.864 &	0.797	& 0.07 \\
Bleeding       (single pattern) &	0.864 &	0.808 &	0.06 \\
Bleeding          (multiple patterns)	& 0.864	& 0.759 &	0.10 \\\hline
\end{tabular}
\caption{\label{tab:tab1}The changes in the direction of the activation vector after augmenting healthy inputs.}
\end{table}
\begin{table}[h]
\centering
\begin{tabular}{|c|c|c|c|}
\hline
Augmented & Average Norm Vector & Average Norm Vector & Average Absolute \\
 Pattern Type &   Original Image &  Augmented Image  & Deviation \\\hline
Fatty dots  (single pattern)&	0.843	&0.792	& 0.05\\
Fatty dots  (multiple patterns) & 0.843	&0.755	&0.09 \\
Cotton wool   (single pattern) &	0.843	&0.803	&0.04 \\
Cotton wool  (multiple patterns) &	0.843 &	0.746	&0.10 \\
\hline
\end{tabular}
\caption{\label{tab:tab2}The changes in the direction of the activation vector after augmenting images having bleeding patterns.}
\end{table}
In addition, similar procedure has been repeated using images with bleeding radiomic pattern (generally from category 2 which have been excluded in our experiments) where other less frequent radiomic patterns have been augmented (Figure \ref{fig:aug2}). The results of this experiment are summarized in Table \ref{tab:tab2}.
The experimental results reveal that increasing the radiomic patterns causes the activation vector to deviate in the direction of the diabetes class. In our experiments, the average deviation of the activation vector is small due to the fact that we augmented the input images with a small number of radiomic patterns, while in the real image of diabetic cases, those patterns occur more frequently. 

\begin{figure}
\centering
\includegraphics[width=.75\linewidth]{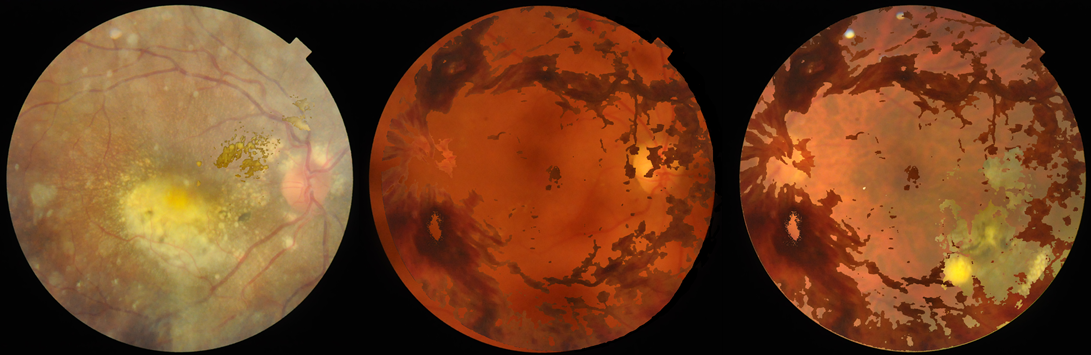}
\caption{\label{fig:exp2}Samples images used in the second experiment.}
\end{figure}

To further verify the effectiveness of the proposed method in determining the contribution of each radiomic pattern to the classifier's decision, we repeated experiments with additional radiomic patterns augmented to the healthy images. We conducted experiments for three different cases where the augmented patterns were fatty dots and cotton-wool spots, bleedings, and a combination of both. Figure \ref{fig:exp2} depicts samples from these three cases.
In addition to measuring the deviation of the concept activation vector in layer n-1, we decided to include the deviations in layer n-2 in this experiment to provide further insights into the behavior of the model in response to the presence of radiomic patterns in input images. The results of this experiment have been summarized in Table \ref{tab:tab3}.
\begin{table}[hbt!]
\centering
\begin{tabular}{|c|c|c|c|}
\hline
Concept & Reference Vector & Angle (degrees) \\\hline
Fatty dots/cotton layer n-1 &	Healthy	& 41\\
Fatty dots/cotton layer n-2 &	Diabetes	& 26\\
Fatty dots/cotton layer n-1 &	Healthy	& 37\\
Fatty dots/cotton layer n-2 &	Diabetes	& 11\\
Bleeding layer n-1 &	Healthy	& 42\\
Bleeding layer n-2 &	Diabetes	& 6\\
Bleeding layer n-1 &	Healthy	& 44\\
Bleeding layer n-2 &	Diabetes	& 6\\
Combined pattern layer n-1 &	Healthy	& 60\\
Combined pattern layer n-2 &	Diabetes	& 5\\
Combined pattern layer n-1 &	Healthy & 56\\
Combined pattern layer n-2 &	Diabetes	& 4\\

\hline
\end{tabular}
\caption{\label{tab:tab3}The measured angles of the concept activation vectors.}
\end{table}
Although we observe deviations in the concept activation vector in both layers n-1 and n-2 for case one of the experiment, the average angle with diabetes cases is relatively high. This result is aligned with the rate of changes in classification results which is 80\%. The deviation in concept activation vector in case two is more consistent as the angle with the angle of diabetes cases has become smaller. These results align with the fact that bleeding is generally the first symptom, and fatty dots or cotton-wool spots rarely occur without the bleeding pattern in an image. In our experiments all images augmented with both fatty dots and cotton-wool, and bleeding patterns have classified as diabetes cases.

\begin{figure}[hbt!]
\centering
\includegraphics[width=.75\linewidth]{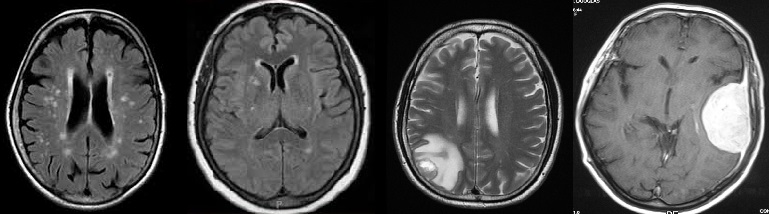}
\caption{\label{fig:mr1}Sample images from the Brain MRI dataset.}
\end{figure}

\begin{figure}[hbt!]
\centering
\includegraphics[width=.75\linewidth]{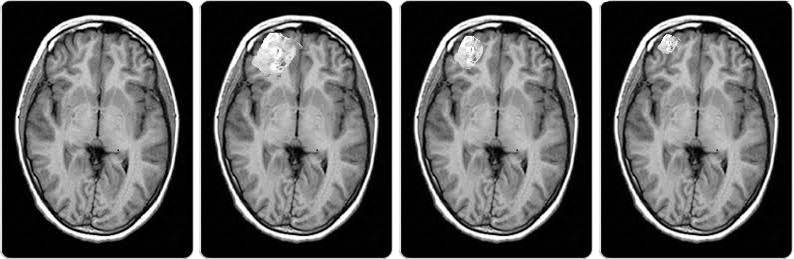}
\caption{\label{fig:mr2}Sample augmented images with different pattern scales from the Brain MRI dataset.}
\end{figure}

 In our second set of experiments we use brain tumor dataset from Kaggle\footnote{https://www.kaggle.com/datasets/navoneel/brain-mri-images-for-brain-tumor-detection}. The dataset consists of 253 brain MRI images, categorized into two groups: images with brain tumors and those without. Each image is provided in JPEG format with varying resolutions. Figure \ref{fig:mr1} presents a selection of sample images from the dataset.  
The experiment conducted by selected an appropriate high level concept. In this experiment we consider the size of the brain tumors as the high level concept. We augmented a subset of healthy images with brain tumors segmented form other images. The segmented brain tumors have been scaled into 3 different categories of large, average, and small. The purpose of the experiment is to determine the impact of the tumor size on the classification decision of the model. Figure \ref{fig:mr2} presents a healthy image augmented with after scaling a segmented tumor. The three images are samples of the test cases.
A CNN model with the same structure as the first experiment is used for the second experiment. The resolution of the input images have been adjusted, as mostly they have much lower resolution than fundus images. The amount of the deviation at activation vectors have been measure and presented in Table \ref{tab:tab4}
\begin{table}[hbt!]
\centering
\begin{tabular}{|c|c|c|c|}
\hline
Concept & Reference Vector & Angle (degrees) \\\hline
Small Brain Tumor  &	Healthy	& 14\\
Small Brain Tumor  &	Diagnosed	& 31\\
Medium Brain Tumor &	Healthy	& 29\\
Medium Brain Tumor &	Diagnosed	& 9\\
Large Brain Tumor  &	Healthy	& 47\\
Large Brain Tumor  &	Diagnosed	& 3\\

\hline
\end{tabular}
\caption{\label{tab:tab4}The measured deviations of the concept activation vectors.}
\end{table}

The experimental results indicate that the trained model is sensitive to the size of the tumors. On average the deviation from the diagnosed with tumor activation vector when a healthy image was augmented with a large scale tumor was insignificant, and all test images were classified as diagnosed with tumor. On the other hand, only a small subset of images augmented with small scale tumors were classified as diagnosed with tumor. In this group the average deviation of the activation vector from healthy class was 14 degrees. The insignificant deviation observed from the healthy activation vector for small tumors in our experiments can be attributed to the training dataset predominantly containing images with large tumors. Consequently, the model has learned to treat large tumor size as a consistent feature. The results of the second experiment demonstrate the applicability of the proposed method across different datasets. However, it is crucial to ensure the presence of the selected high-level concepts during the model's training phase. 

Furthermore, our experiments show that the proposed method has the potential of evaluating the significance of each pattern in isolation. In many medical examinations, the number of determining symptoms is large, and it is of great significance to be able to determine the level of importance the model considers of each of them. In addition, the proposed model facilitates specific consideration about high-level patterns. For instance, the impact of brightness or darkness of radiomic patterns, their texture, size, and location can be effectively verified using the proposed model. \\

In our experiments, we have considered using convolutional neural networks, but the proposed method can be used with any neural network including recurrent networks. 
From computational perspective, the proposed method has the advantage of training and using a single model. Despite the fact that both ACAV and TCAV solutions are model agnostic, however, TCAV requires training a second model which requires not only extra processing, but also a separate dataset of high-level concept. 

Our experimental results were validated by a domain expert, who confirmed that the findings of our proposed method align with expectations and observations from practitioners in hospitals$.^{\ref{sec5}}$ 
\section{Conclusion}
\label{sec4}
In this study, we extended previous work on measuring the impact of high-level patterns on the decisions of ML models by isolating these patterns and eliminating the influence of external factors during the experiments. We proposed a method that can incorporate differences in the shape, color, intensity, location, etc., of high-level patterns and effectively verify their impacts on the classification decisions made by the model. The proposed method not only provides insights into the decision-making process of ML models but also opens the way to incorporating domain expert knowledge into classifiers for more effective and accurate results. Although the experiments conducted were limited to augmenting a small number of high-level radiomic patterns, we demonstrated that the proposed model can successfully isolate the impact of these patterns independently of the surrounding data. This work can be extended by verifying the impact of selecting different locations for augmenting the radiomics. Additionally, the size and intensity/contrast of the radiomic patterns on the model's decision can be investigated further.
\section{Acknowledgement}
\label{sec5}
We are sincerely grateful for the invaluable contributions of Dr. Dilay Ozek, Ophthalmologist at Ankara City Hospital throughout the course of our research. Dr. Ozek's expertise and insights have been instrumental in shaping the direction and depth of our study. Their generous sharing of information, clinical experience, and thoughtful guidance have significantly improved the quality and relevance of our work.

\bibliographystyle{alpha}
\bibliography{main}

\end{document}